\documentclass[letterpaper, 10 pt, conference]{ieeeconf}  % Comment this line out if you need a4paper

\IEEEoverridecommandlockouts                              % This command is only needed if 
\usepackage{graphics} % for pdf, bitmapped graphics files
\usepackage{epsfig} % for postscript graphics files
\usepackage{mathptmx} % assumes new font selection scheme installed
\usepackage{times} % assumes new font selection scheme installed
\usepackage{amsmath} % assumes amsmath package installed
\usepackage{amssymb}  % assumes amsmath package installed
\usepackage{color}
\usepackage{xcolor}
\usepackage{caption}
\usepackage{preambles}
\usepackage{amsmath}
\usepackage{graphicx}
\usepackage{caption}
\usepackage{subcaption}

\title{\LARGE \bf
Robust Feedback Motion Policy Design Using Reinforcement Learning on a 3D Digit Bipedal Robot}
\author{Guillermo A. Castillo$^{1}$, Bowen Weng$^{1}$, Wei Zhang$^{2}$, and Ayonga Hereid$^{3}$% <-this % stops a space
\thanks{*This work was supported in part by the National Science Foundation under grant CNS-1552838, the OSU M\&MS Discovery Theme Initiative, and the startup fund of SUSTech. }% 
\thanks{$^{1}$Electrical and Computer Engineering, Ohio State University, Columbus, OH, USA;  {\tt\footnotesize \{castillomartinez.2, weng.172\}@osu.edu.}}
\thanks{$^{2}$SUSTech Institute of Robotics, Southern University of Science and Technology (SUSTech), China; {\tt\footnotesize zhangw3@sustech.edu.cn.}}
\thanks{$^{3}$Mechanical and Aerospace Engineering, Ohio State University, Columbus, OH, USA. {\tt\footnotesize hereid.1@osu.edu.}}%
}

\usepackage[capitalise]{cleveref}

\usepackage[noadjust]{cite}

\usepackage{graphicx}

\begin{document}
\maketitle
\thispagestyle{empty}
\pagestyle{empty}

\begin{abstract}
In this paper, a hierarchical and robust framework for learning bipedal locomotion is presented and successfully implemented on the 3D biped robot Digit built by Agility Robotics. We propose a cascade-structure controller that combines the learning process with intuitive feedback regulations. This design allows the framework to realize robust and stable walking with a reduced-dimension state and action spaces of the policy, significantly simplifying the design and reducing the sampling efficiency of the learning method. The inclusion of feedback regulation into the framework improves the robustness of the learned walking gait and ensures the success of the sim-to-real transfer of the proposed controller with minimal tuning. 
We specifically present a learning pipeline that considers hardware-feasible initial poses of the robot within the learning process to ensure the initial state of the learning is replicated as close as possible to the initial state of the robot in hardware experiments. Finally, we demonstrate the feasibility of our method by successfully transferring the learned policy in simulation to the Digit robot hardware, realizing sustained walking gaits under external force disturbances and challenging terrains not included during the training process. To the best of our knowledge, this is the first time a learning-based policy is transferred successfully to the Digit robot in hardware experiments without using dynamic randomization or curriculum learning.

\end{abstract}

\section{Introduction} \label{sec:intro}
For the bipedal locomotion problem, the policy \emph{robustness} is a critical characteristic and remains one of the biggest challenges in the field. In practice, the robustness of the control policy can be presented as (i) the capability of handling various external disturbances (e.g., push recovery), (ii) maintaining stable gaits while operating under various terrain conditions, and (iii) accomplishing the sim-to-real transfer with as little effort as possible. In this paper, we present a successful application of designing a feedback motion policy on Digit, a challenging 3D bipedal robot, and demonstrate robust performance among all of the aspects mentioned above. 

\begin{figure}
\centering
\vspace{2mm}
\includegraphics[clip,width=1\columnwidth]{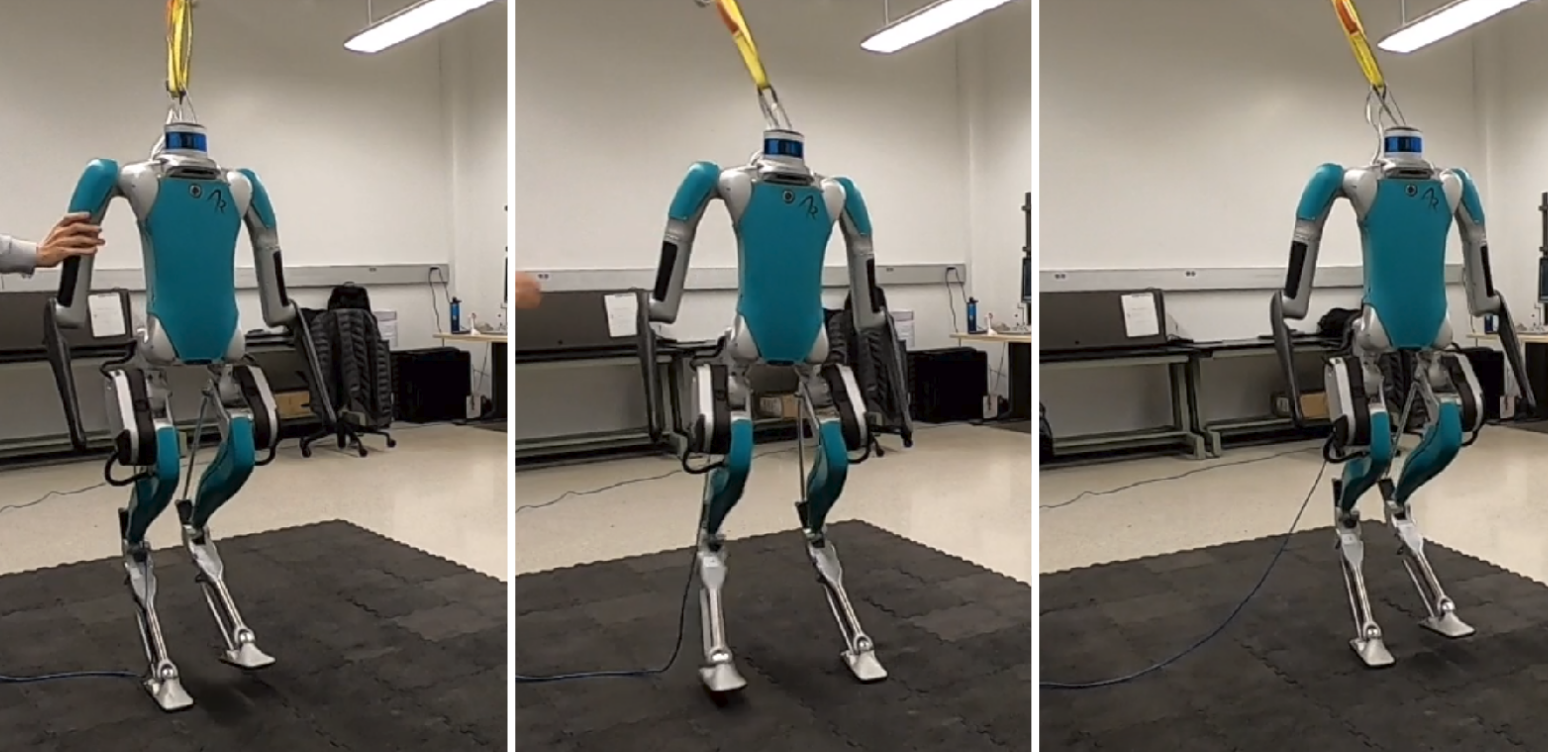}
% \vspace{-7mm}
\caption{Digit recovering from an external force disturbance applied in the lateral direction.} 
\label{fig:disturbancetransition}
% \vspace{-3mm}
\end{figure}

In general, a robust bipedal locomotion policy can be obtained through (i) the model-based approach, (ii) the learning-based approach, or (iii) a combination of both. In the model-based regime, existing research typically relies on a simplified template model~\cite{ashtianirobust2017, chen2020underactuated}. Some has also explored robust control formulations of Control Lyapunov Function Control Barrier Functions (CLF-CBF) \cite{nguyen2020optimal} and Hybrid Zero Dynamics (HZD) inspired solutions \cite{Hereid20163D}. In general, the derived policy from the simplified model requires further tuning of low-level control heuristics. 
On the other hand, a typical learning-based solution involves supervised learning, reinforcement learning, and imitation learning. The policy is typically constructed in an end-to-end manner~\cite{lillicrap2015continuous} directly working with the robot' s full-body dynamics. Some existing work has shown remarkable progress in push-recovery~\cite{yang2018learning} and operating in various terrain conditions~\cite{xie2020allsteps}. While model-based methods often rely on a simplified model and require extensive control gain tuning, most learning-based solutions require a large amount of data.

A locomotion policy combining the learning-based and the model-based solution is typically formulated in a cascade structure. The high-level controller computes a reference trajectory of a selected anchor point (e.g., the center of mass position). The lower-level controller then seeks to track such a learned reference through basic model information such as kinematics. Morimoto et.al.~\cite{morimoto2004simple, morimoto2005poincare} learned the Poincar\'e map of the periodic walking pattern and applied the method to two 2D bipedal robots. Some recent work has proposed to learn the joint-level trajectory for each joint as the reference motion through supervised learning~\cite{da2017supervised} or using reinforcement learning~\cite{Castillo2019Reinforcement, li2019using, Castillo2020Hybrid}. This approach simplifies the design of the lower-level tracking, which can be as simple as a PD controller. 

 With the aforementioned approaches, the empirical successes are mostly evaluated in simulations. Some recent work has seek to tackle the sim-to-real problem through dynamics randomization~\cite{lee2020learning}, system identification \cite{yu2019sim}, and periodic reward composition~\cite{siekmann2020sim}. However, many of these methods still suffer from poor sampling efficiency. Moreover, the robustness is often partially validated emphasizing one of the three mentioned phenomenons. To the best of our knowledge, it remains a challenge to efficiently obtain a locomotion policy that handles external disturbances, uneven terrains, with a little-effort sim-to-real transfer. The challenge further escalates as one considers a 3D, underactuated, and highly nonlinear bipedal system such as the Digit (Figure~\ref{Digit}). %This motivates the studies presented in this paper.
% combine model and data-driven.....

Motivated by the existing challenges, this paper enhances the hybrid zero dynamics inspired reinforcement learning framework presented in our previous work \cite{Castillo2019Reinforcement, Castillo2020Hybrid} emphasizing the policy robustness and sim-to-real transfer. The proposed cascade control structure couples the high-level learning-based controller with the low-level model-based controller. The obtained policy is able to achieve robust bipedal locomotion on a challenging 3D Digit robot in experiments to accomplish the following tasks.
\begin{itemize}
    \item Learn disturbance resistance without explicitly experiencing adversarial attacks in training.
    \item Adapt to various terrain conditions with the policy learning process only limited to flat ground.  
    \item Transfer the learned policy between different simulation environments and from simulation to the real robot without randomized dynamics in training or extensive hyper-parameter tuning.
\end{itemize}
It is worth emphasizing that the policy is learned from scratch and does not rely on demonstrations. To the best of our knowledge, this is the first successful sim-to-real transfer of a learning-based policy working with Digit. We further emphasize the following features that contribute to the aforementioned robust performance of the proposed policy:

\textbf{The high-level learning-based policy} propagates reference motion trajectories in real-time at low-frequency compatible with real hardware implementation. This encourages the stability of the walking gait by modifying the reference trajectories accordingly to the reduced state of the robot.

\textbf{The low-level model-based regulation}
applies compensation to the reference trajectories at a higher frequency based on instantaneous state feedback. This feedback-based trajectory regulation compensates for uncertainty in the environment, and based on our experiment observations, it is one of the key factors leading to the effortless sim-to-real transfer (as illustrated in Fig.~\ref{fig:limit_cycle}).

\textbf{A model-based balancing controller} is implemented that induces a set of feasible initialization states for operations in the simulators and with the real robot. While a feasible initialization is typically not of interest in simulations given the state can be set arbitrarily, it is crucial to the observed success of implementing the proposed locomotion policy to work with Digit in real-world experiments.

The remainder of the paper is organized as follows. Section \ref{sec:model} presents details of the Digit robot. Section \ref{sec:learning} introduces the proposed framework including the high-level learning module, the low-level model-based controllers, the training, and the execution of the overall framework. A series of experiments are included in Section \ref{sec:results} emphasizing the locomotion robustness in terms of disturbance rejection, stable walking gaits on uneven terrains, and effortless sim-to-real transfer. Section \ref{sec:conclusions} concludes the paper and discusses future work.

\section{Kinematic Model of Digit Robot} \label{sec:model}

In this section, we will briefly introduce the Digit robot and its kinematic model and notations that will be used in the following sections.

\subsection{Robot Hardware}
\begin{figure}
\centering
\vspace{2mm}
\includegraphics[clip,height=6cm]{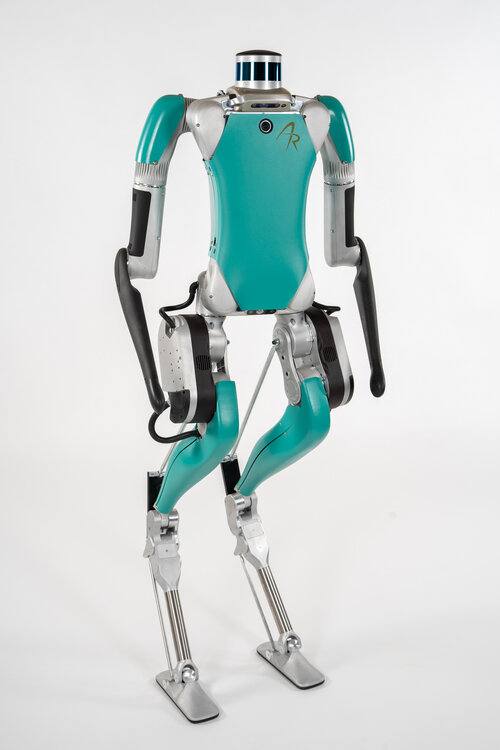}
\includegraphics[clip,height=6cm]{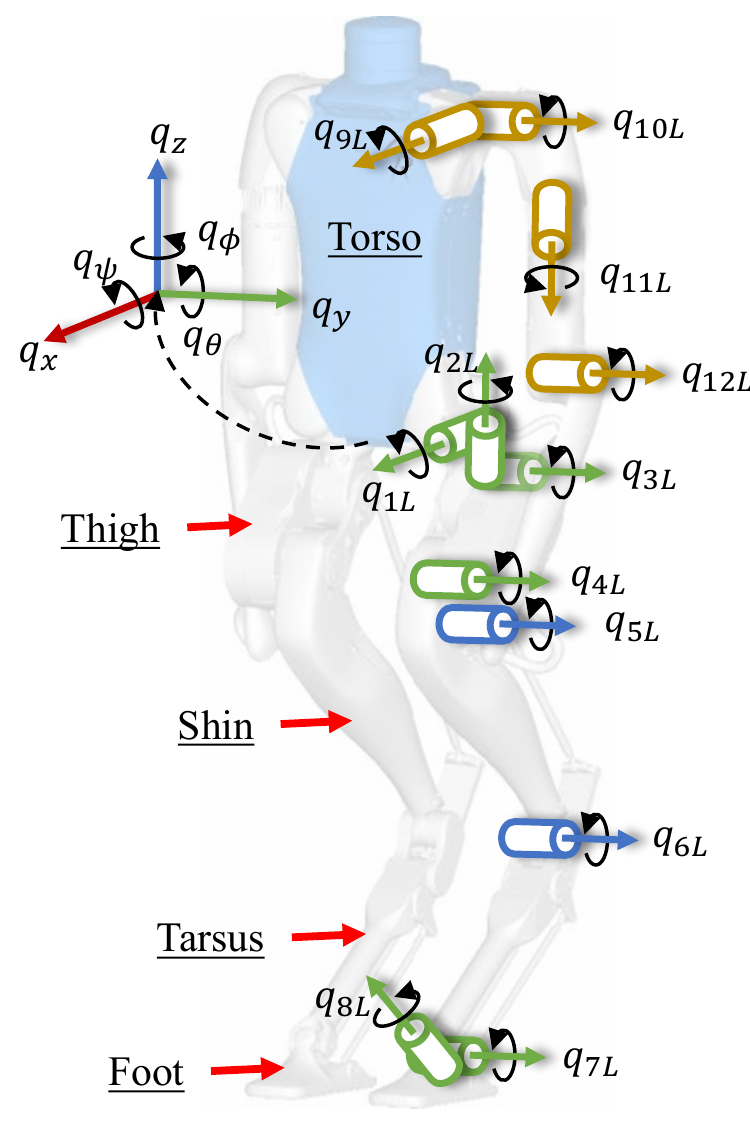}
% \vspace{-7mm}
\caption{Digit robot (left) and its kinematic tree structure (right). The floating base coordinate is located at the pelvis link. Only left leg and left arm joints are shown.  Green represents actuated leg joints, blue represents passive leg joints, and orange represents arm joints. } 
\label{Digit}
\vspace{-3mm}
\end{figure}

Digit is a versatile bipedal robot designed and built by Agility Robotics~\cite{hurst2019building}. The robot has an integrated perception system, 20 actuated joints, 1 IMU, and 30 degrees of freedom (DoF) that allows it to achieve robust and dynamic locomotion. The robot design is based on its predecessor Cassie, whose leg's morphology is derived from an Orchid-like bird. Hence, the location of the knee and ankle is not immediately obvious from the robot's appearance. \figref{Digit}. shows the Digit robot with a description of the body links and its kinematic structure. 
The total weight of the robot is 48 kg. The torso of the robot weighs 15 kg and contains Digit’s on-board computer, power source, and vision sensors. Vision sensors include a LiDAR, an RGB camera, three monochrome depth cameras and a RGB-depth camera.

 %%\figref{Digit}. shows a description of the robot's components.

\subsection{Kinematic Model and Notations}
Each arm of Digit has 4 joints for basic manipulation tasks. Each leg has eight joints, from which four are directly actuated by electrical motors (hip yaw, hip roll, hip pitch, knee), two are indirectly actuated by electrical motors (toeA, toeB) and correspond to the actuated joints of the foot (foot pitch, foot roll). The addition of the foot pitch and foot roll joints improves the ability to balance stably on a wide variety of terrains. The remaining two joints (ankle and shin) are passive and they are connected via specially designed leaf-spring four-bar linkages for additional compliance.

As shown in \figref{Digit}, the variables $q_x, q_y, q_z$ denote the Cartesian position of the robot's pelvis, while the torso orientation is represented by the x-y-z Euler angles (roll, pitch, yaw) as $q_\psi, q_\theta, q_\phi$. Then, the coordinate of the floating base coordinate system at the pelvis is denoted by 
\begin{align}
    \mathbf{q^{b}} = [q_x, q_y, q_z, q_\psi, q_\theta, q_\phi]^T. 
    \label{eq:base}    
\end{align}
The actuated joints are denoted by the vector given as,
\begin{align}
    \mathbf{q^{a}} = [q_{1L}, q_{2L}, q_{3L}, q_{4L}, q_{7L}, q_{8L},q_{9L},q_{10L},q_{11L},q_{12L},\notag      \\
    q_{1L}, q_{2L}, q_{3L}, q_{4L}, q_{7L}, q_{8L},q_{9L},q_{10L},q_{11L},q_{12L}]^T, \label{eq:actjoints}
\end{align}
where the joints $q_1$-$q_4$ correspond to the leg's actuated joints (i.e. hip roll, hip yaw, hip pitch, knee), joints $q_7, q_8$ correspond to the foot pitch and foot roll, and joints $q_9$-$q_{12}$ correspond to the arm's actuated joints (i.e. shoulder roll, shoulder pitch, shoulder yaw, elbow).
The passive joints are denoted by the vector:
\begin{align}
    \mathbf{q^{p}} = [q_{5L},q_{6L}, q_{5R},q_{6R}]^T , \label{eq:pasjoint}
\end{align}
where $q_5, q_6$ correspond to the shin and tarsus joints. 
Hence, the generalized coordinates of Digit robot are denoted by:
\begin{align}
    \mathbf{q} = [\mathbf{q^{b}}, \mathbf{q^{a}}, \mathbf{q^{p}}]^T. 
    \label{eq:generalized coordinates}    
\end{align}

In addition, the robot has 4 end-effectors, which are the left/right feet and fists. We define the Cartesian position of any of these end-effectors as:
\begin{align}
    \mathbf{x}_{ee}\mathbf{(q)} = [x_{ee}, y_{ee}, z_{ee}]^T 
    \label{eq:end-effector}    
\end{align}

The CoM Cartesian position is denoted by:
\begin{align}
    \mathbf{x}_{CoM}\mathbf{(q)} = [x_{CoM}, y_{CoM}, z_{CoM}]^T 
    \label{eq:CoM}    
\end{align}

The current position of the robot's feet and CoM can be determined by applying Forward Kinematics (FK). In particular, we use the the open-source package FROST \cite{Hereid2017FROST} to obtain symbolic expressions for $\mathbf{x}_{footL}\mathbf{(q)}$, $\mathbf{x}_{footR}\mathbf{(q)}$, and $\mathbf{x}_{CoM}\mathbf{(q)}$. We created a URDF model of Digit based on the XML model provided by Agility Robotics.

\section{Learning Approach} \label{sec:learning}
In this section, we build upon our previous work proposed in \cite{Castillo2019Reinforcement, Castillo2020Hybrid} to implement a cascade-structure learning framework that realizes stable and robust walking gaits for the 3D bipedal robots. The specific design ensures successful transferring learned policies in simulation to robot hardware with minimal turning.

% that can be easily transferred to real hardware with minimal tuning. 

\begin{figure}
\vspace{2mm}
\centering
\includegraphics[trim={0cm 0cm 0cm 0cm},clip,width=1\columnwidth]{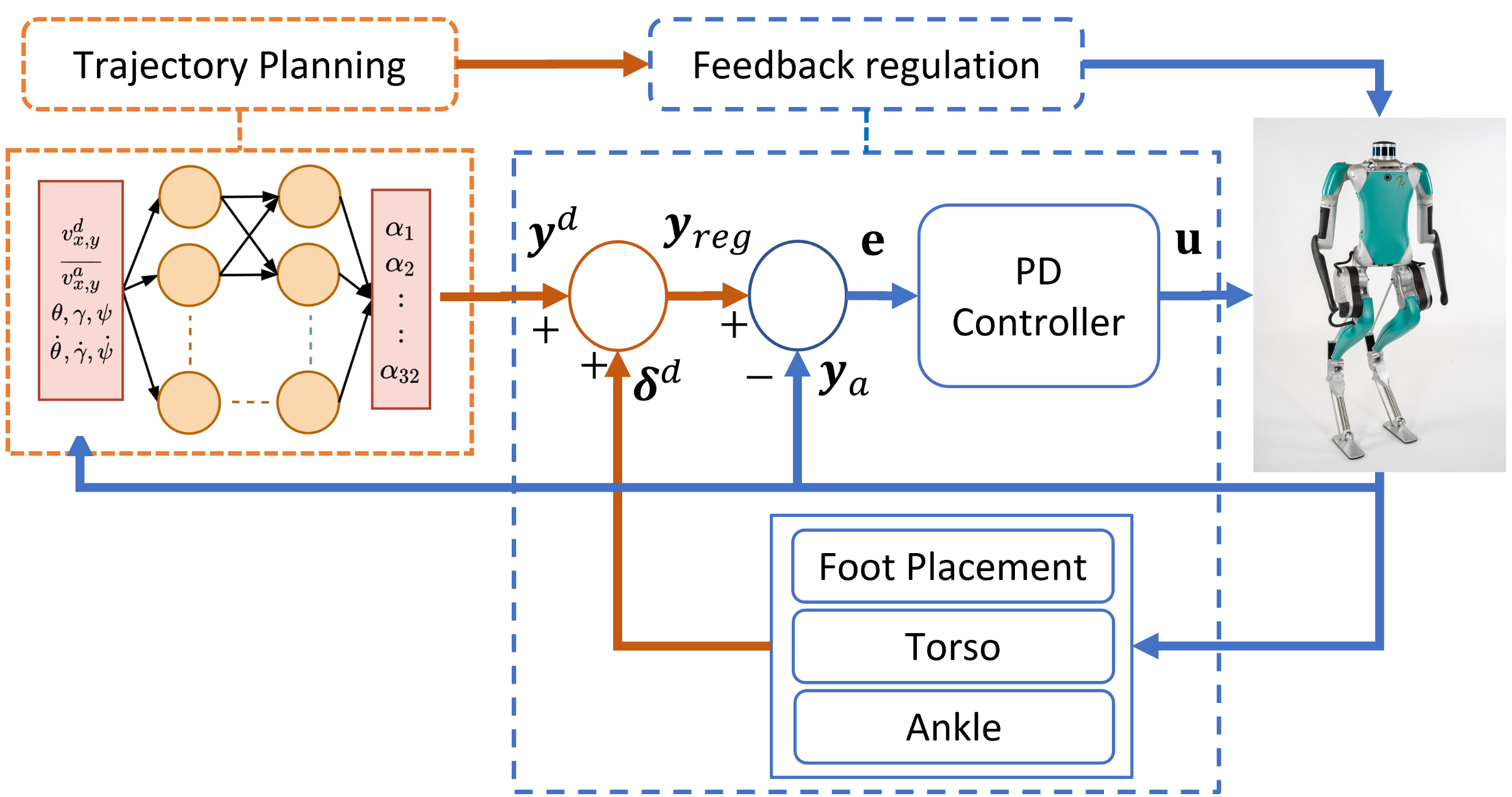}
\caption{Overall structure of the proposed Trajectory-based RL framework. The trajectory planning phase is done by the neural network policy while the feedback regulation block uses the robot's sensor information to improve the stability of the walking gait and the velocity tracking performance. }
\label{fig:overall_structure}
\vspace{0mm}
\end{figure}

% This structure is shown in Fig. ..... where we can clearly differentiate the two stages of the framework. 

\subsection{Overview of the Learning Framework}
As shown in \figref{fig:overall_structure}, the proposed framework presents a cascade decoupled structure to tackle the 3D walking problem through its two main components, neural-network trajectory planning and feedback regulation.

The neural network trajectory planner uses a reduced-dimensional representation of the robot states to compute a set of coefficients $\mathbf{\alpha}$ that parameterize joint reference trajectories using $5^{th}$-order Bézier Polynomials. The structure of the neural network is provided in \secref{learning_NN}. Since the forward and backward propagation  of the neural network could take significant time when compared with the running frequency of the low-level controller (1 kHz), the trajectory planning phase runs at a lower frequency (250 Hz). This consideration is especially relevant for the implementation of the learned policy in the real robot. More importantly, it opens the door to the real-time implementation of new learning structures that may be computationally expensive in both simulation and hardware. 

The feedback regulation uses the robot's pelvis velocities and torso orientation to modify the joint reference trajectories. It compensates for the uncertainty in the robot's model and the environment, improving the robustness of the walking gait. The joint-level PD controllers are used to track the compensated reference joint trajectories. The compensations applied during the feedback regulation are discussed in detail in \secref{sec:regulations}. The low-level feedback regulation is a key component of our control structure to close the sim-to-real gap, enabling successful transfer of the learned policy to hardware without the need for exhaustive tuning, dynamic randomization, or curriculum learning.  

% As mentioned before, the low-level feedback regulation runs at a higher frequency than the motion planning, which is 

\subsection{Dimension Reduction of the State and Action Space}\label{dimension_reduction}
By applying physical insights from the walking motion, we can significantly reduce the number of inputs and outputs of the NN. Unlike other end-to-end RL frameworks, we do not use all the available states of the robot to feed the NN. Instead, we select states that provide insightful information such as the torso orientation ($q_\psi, q_\theta, q_\phi$), pelvis linear velocity ($\dot q_x, \dot q_y, \dot q_z$), torso angular velocity ($\dot q_\psi, \dot q_\theta, \dot q_\phi$), and desired walking direction, which is encoded by the desired forward and lateral walking velocity ($\dot q_x^d, \dot q_y^d$).

To reduce the dimension of the action space, we will keep the arms joints fixed during the walking motion and keep the stance foot passive and the swing foot parallel to the ground during the walking gait. Without the need for determining reference trajectories for arm and foot joints, we reduce the number of reference trajectories needed to be computed by the NN to eight, with each leg having three hip joints and one knee joint. In addition, we impose a symmetry condition between the right and left stance during the walking gait. Therefore, given the set of coefficients for the right stance $\mathbf{\alpha}^R \in \mathbb{R}^{M+1\times N}$, where $M=5$ is the degree of the Bézier polynomials and $N=8$ is the number of reference trajectories, we can obtain the set of coefficients for the left stance $\mathbf{\alpha}_L \in \mathbb{R}^{M+1\times N}$ by the symmetry condition:
\begin{align}
    \label{eq:symm_cond}
    \mathbf{\alpha}^L =  \mathbf{T} \mathbf{\alpha}^R,
\end{align}
where $\mathbf{T} \in \mathbb{R}^{N \times N}$ is a sparse transformation matrix that represents the symmetry between the joints of the right and left legs of the robot.

To further reduce the action space and encourage the smoothness of the control actions after the ground impact, we enforce that at the beginning of every step, the initial point of the B\'ezier polynomial coincides with the current position of the robot's joints. That is, for each joint $i$ with Bézier coefficients $\mathbf{\alpha}^R_i \in \mathbb{R}^{M+1}$, we have
\begin{align}
    \label{eq:impact_cond}
    \alpha_i[0] = q_i(\tau(0)), \quad \tau \in [0,1],
\end{align}
where $\tau$ is the time-based phase variable used to parameterize the Bézier Polynomials.
Finally, we enforce the position of the hip joints and knee joints to be the same at the end of the current step and the beginning of the next step ($\tau(t)=0$ and $\tau(t)=1$ respectively). This enforces continuity in the joint position trajectories after switching the stance foot. 

\subsection{Neural Network Structure} \label{learning_NN}
Given the considerations presented in \secref{dimension_reduction}, the number of inputs for the NN is $10$ and the number of outputs is $32$. We choose the number of hidden layers of the NN to be $4$, each one with $32$ neurons. The activation function for the hidden layers is ReLU, and the activation function for the output layer is sigmoid. Finally, we scale the output of the NN within a range of admissible motion for the robot's joints. In particular, we use the convex hull property of Bézier polynomials to translate the joint limits to the corresponding coefficients limits. Given the number of inputs and outputs of the NN, the number of trainable parameters is 4576. To the best of our knowledge, this is the smallest NN implemented in hardware to realize 3D walking locomotion.

\subsection{Walking Policy Learning Pipeline} \label{learning_procedure}

\begin{figure}
\centering
\vspace{2mm}
\includegraphics[clip,width=1.0\columnwidth]{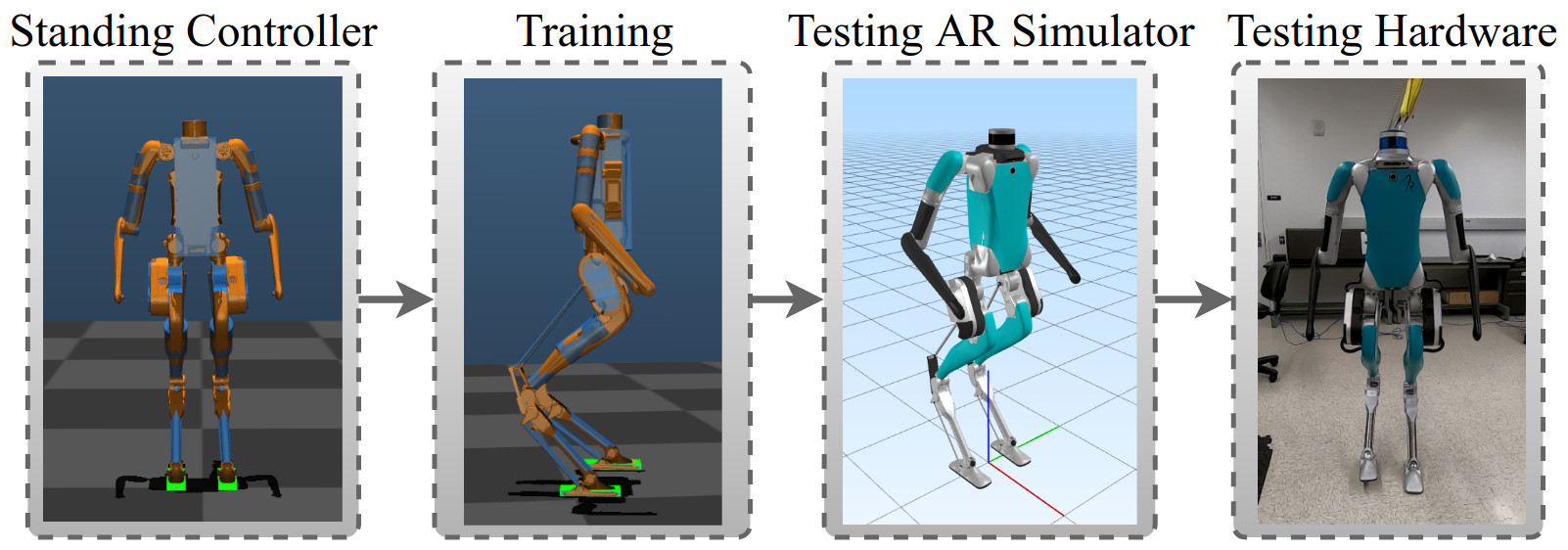}
% \vspace{-7mm}
\caption{Pipeline of the proposed learning framework. Standing controller is included within the pipeline of the learning process to obtain a feasible set of initial states for the policy training using a customized Mujoco environment. Then, the trained policy is tested in a more realistic real-time simulator, and finally it is tested on hardware. } 
\label{fig:pipeline}
% \vspace{-3mm}
\end{figure}

Since the main purpose of this work is to implement a learning framework that realizes a policy that is transferable to the real hardware, we need to create a pipeline for the training process in simulation that renders working conditions as close as possible to the real hardware. A diagram of the training pipeline is shown if \figref{fig:pipeline}.

The initial state for each training episode is chosen randomly from a pool of initial states whose kinematics and dynamics are feasible to be implemented on the real robot. To achieve this, we implement a standing controller that allows the robot to start up from an arbitrary resting position when the robot is hanging up from the lab's crane. We tested the controller successfully in hardware and use it to replicate the standing process in simulation. More details about the standing controller are provided in \secref{standing}.

After the standing controller is activated and the robot is able to stand and balance steadily in simulation, we capture the state of the robot and save it in a pool of initial states. Then, we repeated the process 40 times to create a diverse enough set of initial states that we can use for training our controller. It is important to denote that this process does not provide a fixed initial state for the walking gait, but a whole set of different initial configurations that encourage the randomness of the initial state during the training while providing feasible and safe starting conditions for the walking gait. Moreover, this process does not provide any reference trajectories for the walking gait that could bias the learning process to specific or predefined walking gaits. That is, different from other RL approaches to bipedal locomotion, our method learns walking gaits from scratch without the need for predefined reference trajectories or imitation learning. 

Given the pool of initial states, we use a customized environment using MuJoCo \cite{Todorov2012MuJoCo} to start the training process using the Evolution Strategies (ES) algorithm. When the training is finished, we proceed to test the trained policy in the Agility Robotics proprietary simulation software. This simulation software provides a more realistic representation of the robot's dynamic behavior since it runs in real-time and shares the same features with the hardware such as low-level API, communication delay, and model dynamic parameters. 

Finally, once we verify the trained policy is deployed safely in the AR simulation, we proceed to test it in the hardware and verify its performance to do additional tuning of the low-level controller gains if needed. A sequence of the whole learning process and transference to real hardware can be seen in the accompanying video submission. %\cite{video_link}

\subsection{Reward Function Design} \label{learning_reward}
Following the work in \cite{Castillo2020Hybrid}, we use a reward function of the form: 
% \begin{equation}
%     \label{eq:reward func}
%     r = \mathbf{w}^T \mathbf{r},
% \end{equation}
% with a vector of nine customized rewards $\mathbf{r}$ and the weights $\mathbf{w}$. Specifically, 
\begin{equation}
    \mathbf{r} = \mathbf{w}^T [ r_{v_x}, r_{v_y}, r_h, r_a, r_{CoM}, r_{ang}, r_{angvel}, r_{u}, r_{fd} ]^T.
\end{equation}
where $\mathbf{w}$ is a vector of weights corresponding to each of customized rewards $r_i$. 
This reward function encourages forward and lateral velocity tracking (through $r_{v_x}, r_{v_y}$), height maintenance (with $r_h$), energy efficiency (with $r_{u}$) and natural walking gaits. 
The episode length is 10000 simulation steps, which are equivalent to 5 seconds, and it has an early termination if any of the following conditions is violated: 
\begin{equation}
    \begin{aligned}
    &|q_\psi|<0.5, \quad |q_\theta|<0.5, \quad |q_\phi|<0.5, \\ 
    &|\dot{q}_\psi|<2, \quad |\dot{q}_\theta|<2, \quad |\dot{q}_\phi|<2, \\ 
    &0.8<q_z<1.2, \quad \Delta_f < 0.05,
    \end{aligned}
\end{equation}
where $q_z$ is the height of the robot's pelvis and $\Delta_f$ is the distance between the feet.

\section{Control} \label{sec:control}
\begin{figure}
\centering
\vspace{2mm}
\includegraphics[clip,width=1\columnwidth]{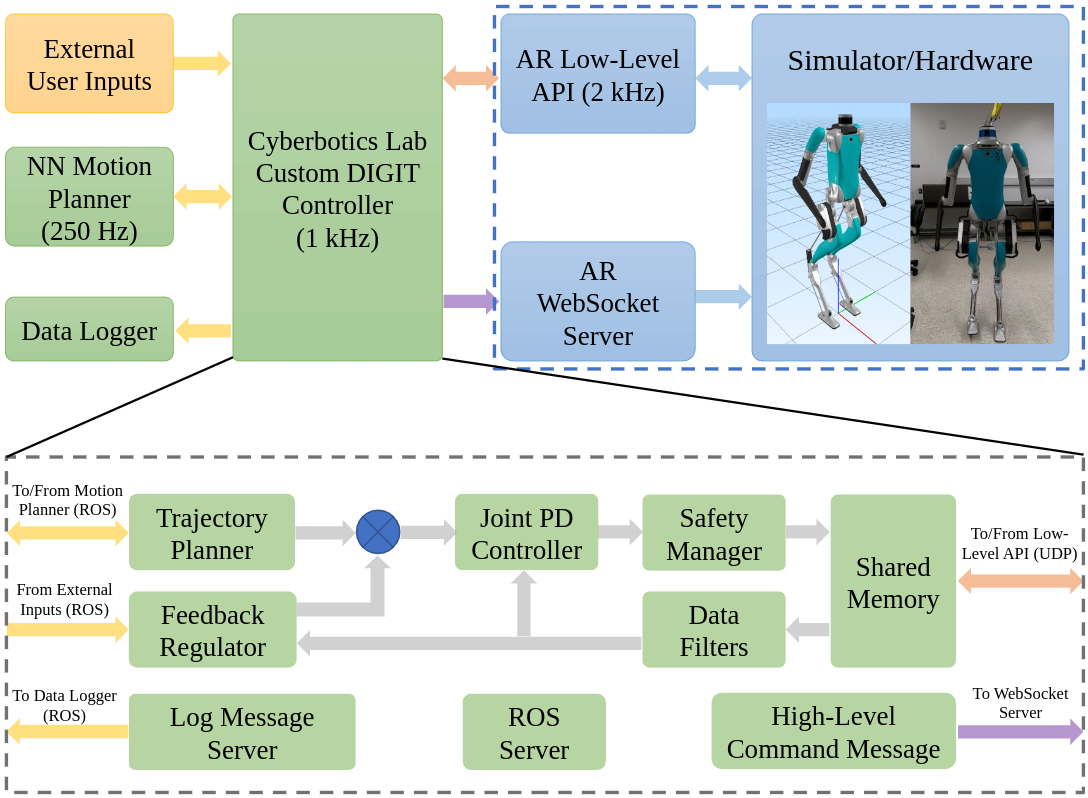}
% \vspace{-7mm}
\caption{Our controller architecture allows asynchronous operation of the high-level and low-level controllers. Moreover, it enables fast and reliable communication between different control layers, external user inputs, and the Digit simulator/hardware. Note that the same controller architecture can be used for different tasks e.g. walking, standing, crouching.} 
\label{fig:controller_architecture}
% \vspace{-3mm}
\end{figure}

In this section, we provide details about the implementation of the high-level and low-level controllers of the control-learning framework introduced in \secref{sec:learning}. Moreover, we provide details of the architecture of our controller and its integration with Digit's low-level API and communication system. This architecture is presented in \figref{fig:controller_architecture}. 

The Agility Robotics low-level API streams data using the UDP protocol and enables the user to access the low-level sensor data and to give commands directly to the motor drives. To keep the connection of the low-level API active to receive sensor data, commands must be sent periodically. Once the client is connected and sending commands, torque control must be activated by requesting transition to Low-level API operation using JSON messages sent through Websocket protocol. 

In addition, we use the Robot Operating System (ROS) to manage the communication between different components in our system. This also adds significant flexibility to our controller structure to include nodes for additional tasks such as logging information, capturing external command inputs, and accessing Digit's perception system.

Since the low-level API needs to run at a very high frequency (2 kHz), we use shared memory to enable fast communication between the low-level API and our custom main control code (1 kHz). The low-level API reads the sensor measurements and writes them into the shared memory. It also reads from the shared memory the torque information and velocity commands written by the main control code and sends the commands to the motors. 

On the other hand, the main control code manages the integration and synchronization of all the components in the overall control structure. It reads the sensor information from the shared memory and publishes it in ROS topics to make it available for the other components of the systems such as the high level planner. In this work, the high-level planner is the trained NN policy, which takes the information from the corresponding ROS topics to compute the output (coefficients of the Bézier Polynomials) at a frequency of 250Hz. Then, the high-level planner publishes this information in its respective ROS topics to make it available for the main control code. 

The main control code reads the Bézier coefficients published by the high-level planner and uses them to compute the corresponding Bézier Polynomials that become the reference trajectories for the robot's joints. In addition, the main control code uses the sensor feedback information shared by the Low-level API to compute the regulations needed to compensate for the model mismatch between the simulation and hardware. The integration of these Low-level feedback regulations into the learning framework is a key part that makes the controller structure very robust to uncertainties in the model and makes possible an almost zero effort transfer from the policy learned in simulation into the real hardware. The detailed structure of the regulations used in the low-level controller is  discussed in \secref{sec:regulations}.

% Finally, after applying the trajectory regulations and tracking of reference trajectories, the main control code writes into the shared memory the commanded torque and velocity for all the robot's joints. 

\subsection{Standing Controller} \label{standing}
The standing controller implemented for Digit is based on the standing controller implemented for Cassie in \cite{Gong2019Feedback}. To ensure the robot keeps the balance, the standing controller uses simple and intuitive regulations based on the torso orientation and the position of the CoM with respect to the feet. That is, keeping the robot CoM within the support polygon while both feet are flat on the ground. The CoM position in the y-axis is adjusted by varying the legs' length, and the CoM in the x-axis is regulated by controlling the pitch angle of the feet.

In our learning framework, the standing controller is used to bring the robot to a stable configuration after initialization in arbitrary standing positions. This process is repeated several times in simulation until we obtain a diverse enough pool of stable configurations that are realistic and feasible to be implemented in hardware. This set of feasible initial configurations is then used to randomly choose the initial states during the training process.

\subsection{Feedback Regulations} \label{sec:regulations}

% One of the contributions of this work is a general framework that allows to transfer policies learned in simulation into real hardware. To this end, 
The feedback regulation module in our controller structure is a key component of the framework as it allows the controller to compensate the trajectories obtained from the high level planner to adapt it to unknown disturbances such as the mismatch between the simulation model and real robot, and environmental factors like external disturbances or challenging terrains that the policy has not experimented in simulation. These regulations are separated in three main groups: foot placement, torso, and foot regulations.

\newsec{Foot placement regulation} controller has been widely used in 3D bipedal walking robots with the objective of improving the speed tracking and the stability and robustness of the walking gait~\cite{da20162d, rezazadeh2015spring, Gong2019Feedback}. Longitudinal speed regulation, defined by \eqref{eq:long_reg}, sets a target offset in the swing hip pitch joint, whereas lateral speed regulation \eqref{eq:lat_reg} do the same for the swing hip roll angle. Direction regulation \eqref{eq:yaw_reg} add an offset to the yaw hip angle to keep the torso yaw orientation at the desired angle.
\begin{align}
    \label{eq:lat_reg}
    \delta_{q_{1i}} &= S_y(\tau(K_{p_{y}}(\dot{q}_y-\dot{q}_y^d) + K_{d_{y}}(\dot{q}_y-\dot{q}_y^{ls})) + \beta_y),\\
    \label{eq:yaw_reg}
    \delta_{q_{2i}} &= \tau (q_\phi - q_\phi^d), \\
    \label{eq:long_reg}
    \delta_{q_{3i}} &= -(\tau(K_{p_{x}}(\dot{q}_x-\dot{q}_x^d) + K_{d_{x}}(\dot{q}_x-\dot{q}_x^{ls})) + \beta_x),
\end{align}
where $i \in \{L,R\}$ depends on which foot is the swing foot, $S_y = 1$ if $i=L, S_y = -1 $ if $ i=R$, $\dot{q}_x$, $\dot{q}_y$ are the longitudinal and lateral speeds of the robot, $\dot{q}_x^{ls}$, $\dot{q}_y^{ls}$ are the speeds at the end of the previous step, $\dot{q}_x^d$, $\dot{q}_y^d$ are the reference speeds, and $K_{p_{x}}, K_{d_{x}}, K_{p_{y}}, K_{d_{y}}$ are the proportional and derivative gains.

\begin{figure*}
\vspace{2mm}
\centering
\begin{subfigure}{.49\textwidth}
  \centering
  \includegraphics[clip,width=\linewidth]{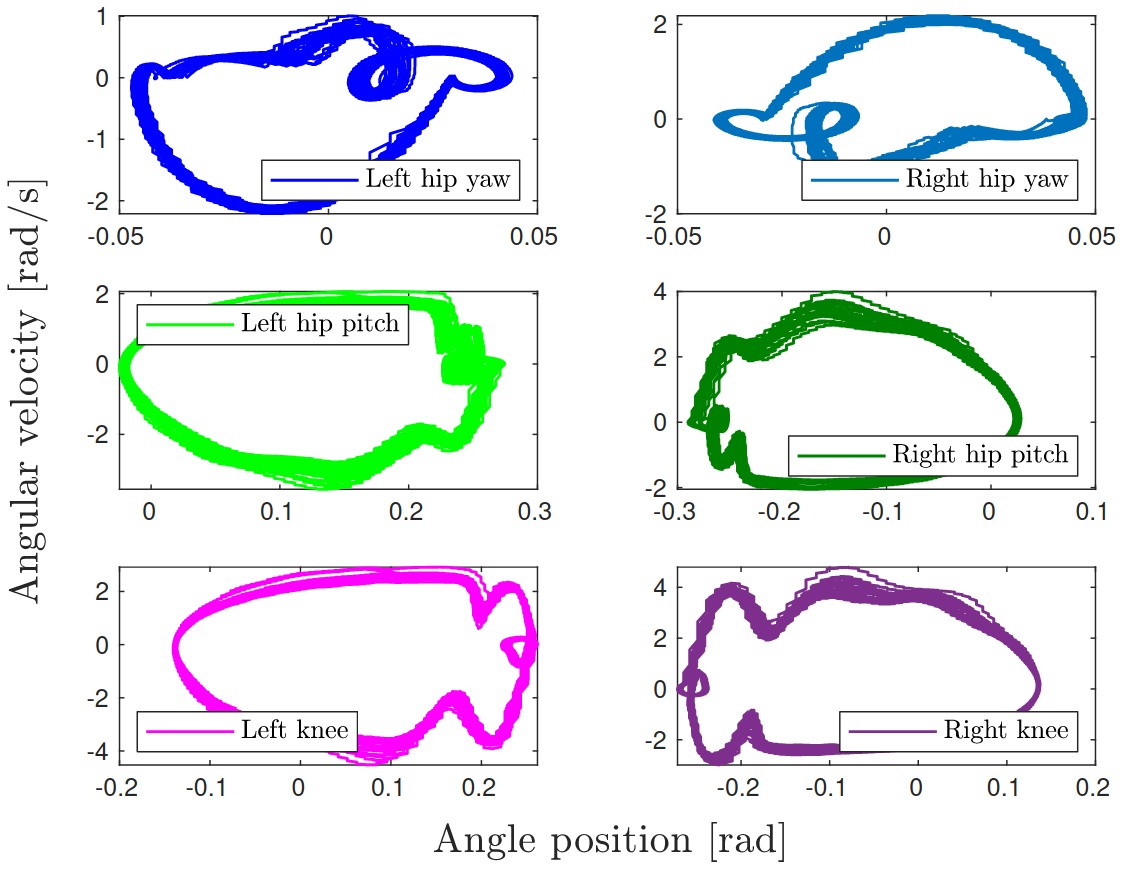}
  \caption{Limit walking cycle of trained policy in simulation. } 
    \label{fig:sub_limit_cycle_sim}
\end{subfigure}%
\begin{subfigure}{.48\textwidth}
  \centering
  \includegraphics[clip,width=\linewidth]{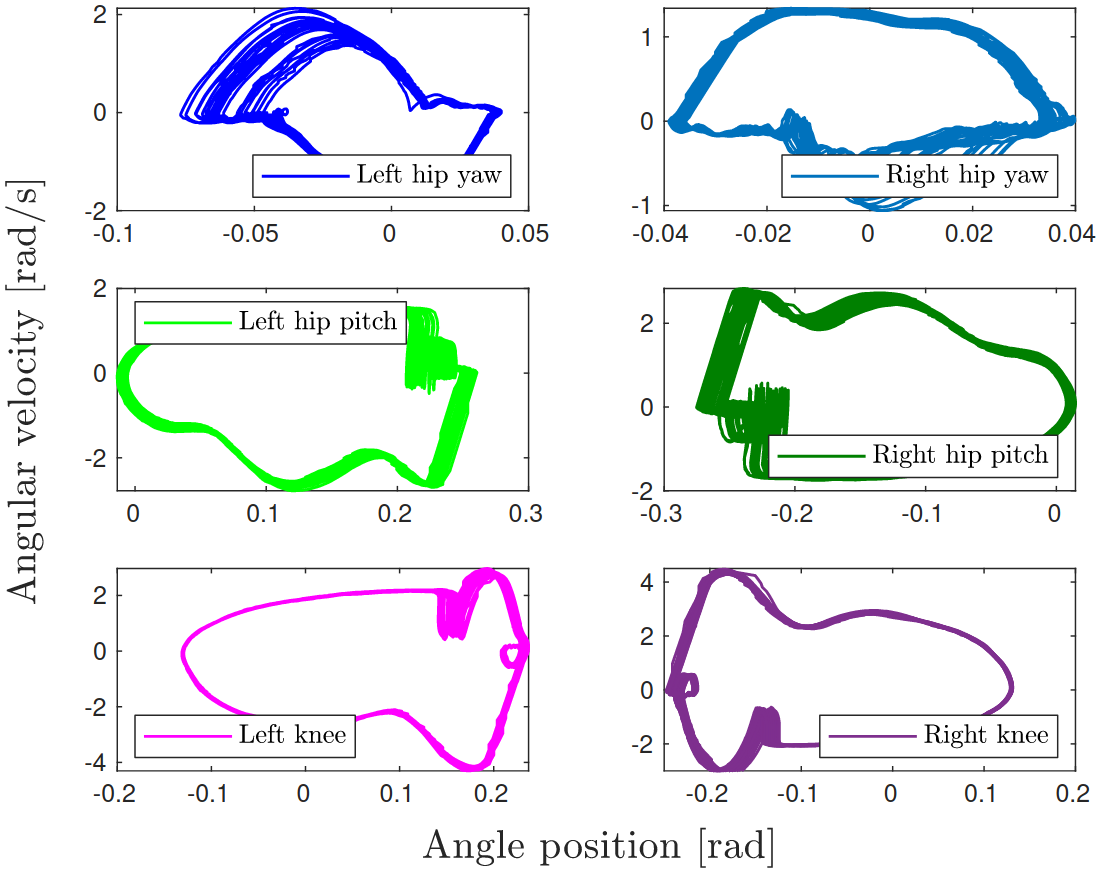}
  \caption{Limit walking cycle of trained policy in real robot. } 
    \label{fig:sub_limit_cycle_real}
\end{subfigure}
\caption{Comparing limit walking cycles between the simulation trails and real-robot tests: (i) For the same joint, simulated motion and real robot motion share similar torque limits. (ii) For all of the hip joints, the regulation term modifies the actual motion leading to the slight difference of the limit walking cycles. (iii) As the regulation is not involved in any of the knee joints, the limit walking cycles are almost identical.}
\label{fig:limit_cycle}
\end{figure*}

The phase variable $\tau$ is used to smooth the regulation at the beginning of each walking step and reduce torque overshoots. The term $\beta$ is the output of an additional PI controller used to compensate for the accumulated error in the velocity and prevent the robot to drift towards a non-desired direction. Based in our experiments, the inclusion of $\beta$ is key in the successful sim-to-real transfer of our controller.

\newsec{Torso regulation} is applied to keep the torso in an upright position, which is desired for a stable walking gait. Assuming that the robot has a rigid body torso, simple PD controllers defined by \eqref{eq:torso_roll} and \eqref{eq:torso_pitch} can be applied respectively to the hip roll $q_{1i}$ and hip pitch $q_{3i}$ angle of the stance leg:
 \begin{align}
    \label{eq:torso_roll}
    u_{q_{1i}} &= -(K_{p_{troll}}(q_\psi - q_\psi^{d}) + K_{d_{troll}}(\dot{q}_{\psi} - \dot{q}_{\psi}^{d})),\\
    \label{eq:torso_pitch}
    u_{q_{3i}} &= S_\theta (K_{p_{tpitch}}(q_\theta - q_\theta^{d}) + K_{d_{tpitch}}(\dot{q}_\theta - \dot{q}_{\theta}^{d})),
\end{align}
where $i \in \{L,R\}$ depends on which foot is the stance foot, $S_\theta = 1$ if $i=L, S_\theta = -1 $ if $ i=R$
where $q_\phi$ and $q_\theta$ are the torso roll and pitch angles, and $K_{p_{troll}}, K_{d_{troll}}, K_{p_{tpitch}}, K_{d_{tpitch}}$ are manually tuned gains.

\newsec{Foot orientation regulation} is applied to keep the swing foot flat during the swinging phase to ensure a proper landing of the foot on the ground. Since Digit has 2 DoF actuated foot, one regulation is needed for each of the roll and pitch angles. By using forward kinematics, the swing foot roll and pitch regulation are determined by 
  \begin{align}
    \label{eq:flat_foot_reg}
    q_{7i}^d &= q_\psi + q_{1i} + S_f (21\deg) \\
    q_{8i}^d  &= q_\theta +q_{3i} + S_f (6\deg), 
\end{align}
where $i \in \{L,R\}$ depends on which foot is the swing foot, $S_f = 1$ if $i=L, S_f = -1 $ if $ i=R$, and $q_{7i}^d, q_{8i}^d$ are the desired pitch and roll foot angle. 
Moreover, the stance foot is kept passive during the stance phase, which helps to the stability of the walking gait, especially for soft or irregular surfaces. 

% It is important to denote that the speed and torso regulations presented above are fixed, intuitive, and applicable to any general 3D bipedal walking robot. However, given the decoupled structure of the controller used for the different regulations, there are several gains that need to be manually tuned in order to achieve improved stability, which is time-consuming and requires experience. However, this process can be easily automated within an RL framework. 

% \input{sections/Sec5_Results}
\section{Experimental Results} \label{sec:results}
In this section, we validate the learning framework and controller structure presented in this paper through both simulation and real hardware experiments on the Digit robot. We prove that our controller structure allows the transference of walking policies learned in simulation to  the real hardware with minimal tuning and enhanced robustness. To the best of our knowledge this is the first time that a policy learned in simulation is successfully transferred to hardware to realize stable and robust dynamic walking gaits for Digit robot. In addition, we show that the improved controller structure presented in this work is robust enough to mitigate the uncertainty in the robot's dynamics caused by the mismatch between the simulation and real hardware. Finally, by thorough experiments on hardware, we show the robustness of the learned controller against adversarial disturbances applied to the robot as well as its capability to adapt to various terrains without need of training for such challenging scenarios. 

\subsection{Sim-to-real Transfer and Stability of the Walking Gait}
We transferred the learned policy trained in simulation into the real robot and we tested empirically the stability of the walking gait by analysing the walking limit cycle described by the robot's joint during the walking motion. \figref{fig:sub_limit_cycle_sim} shows the limit cycle described by some of the robot's joints while the robot is walking in place during about 1 minute in simulation, while \figref{fig:sub_limit_cycle_real} shows the limit cycles from the real experiment.  The convergence of the walking limit cycle to a stable periodic orbit demonstrates that the walking gait is stable and symmetric, meaning that the policy is transferred successfully from simulation to hardware.

\subsection{Robustness to Disturbances}
\begin{figure}
\centering
% \vspace{2mm}
% \includegraphics[trim={0cm 2.2cm 0cm 2.2cm},clip,width=0.9\columnwidth]{figures/Digit_v3_standing.jpg}
\includegraphics[clip,width=0.9\columnwidth]{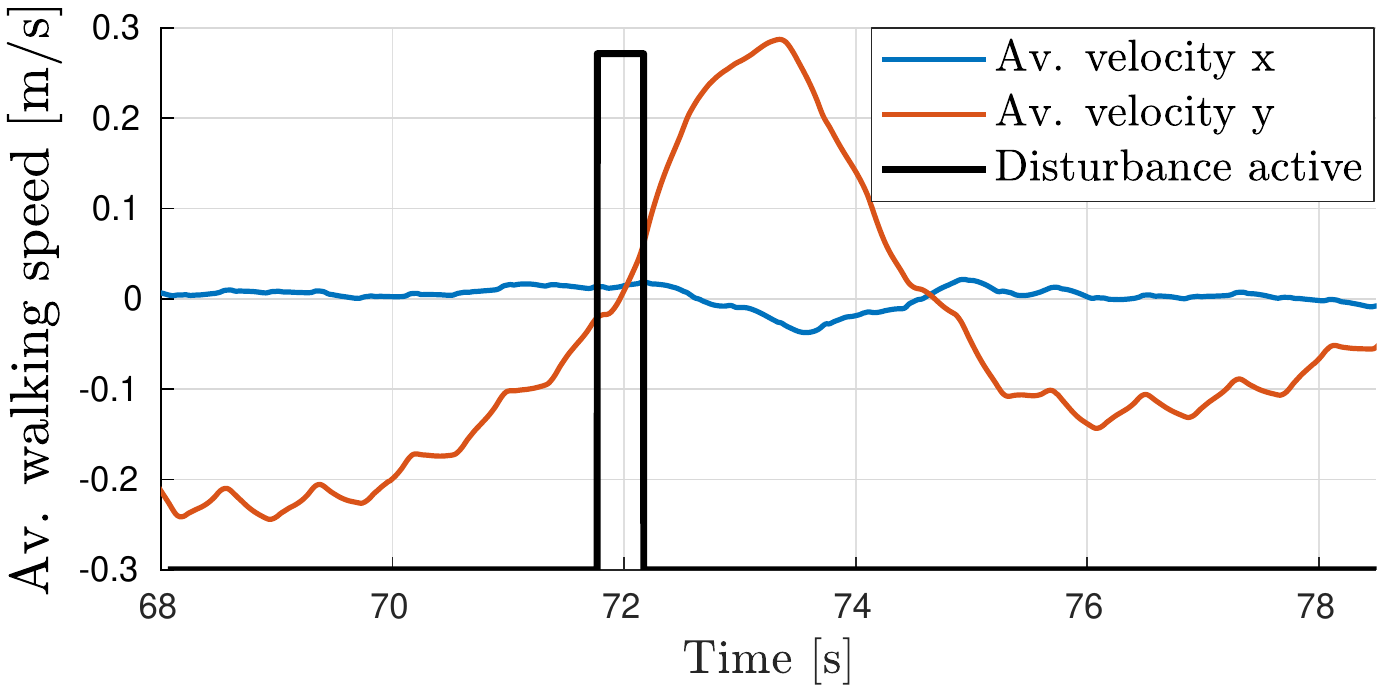}
% \vspace{-7mm}
\caption{Digit recovering from an external disturbance} 
\label{fig:disturbance_vel}
% \vspace{-3mm}
\end{figure}

\begin{figure}
\centering
% \vspace{2mm}
% \includegraphics[trim={0cm 2.2cm 0cm 2.2cm},clip,width=0.9\columnwidth]{figures/Digit_v3_standing.jpg}
\includegraphics[clip,width=1\columnwidth]{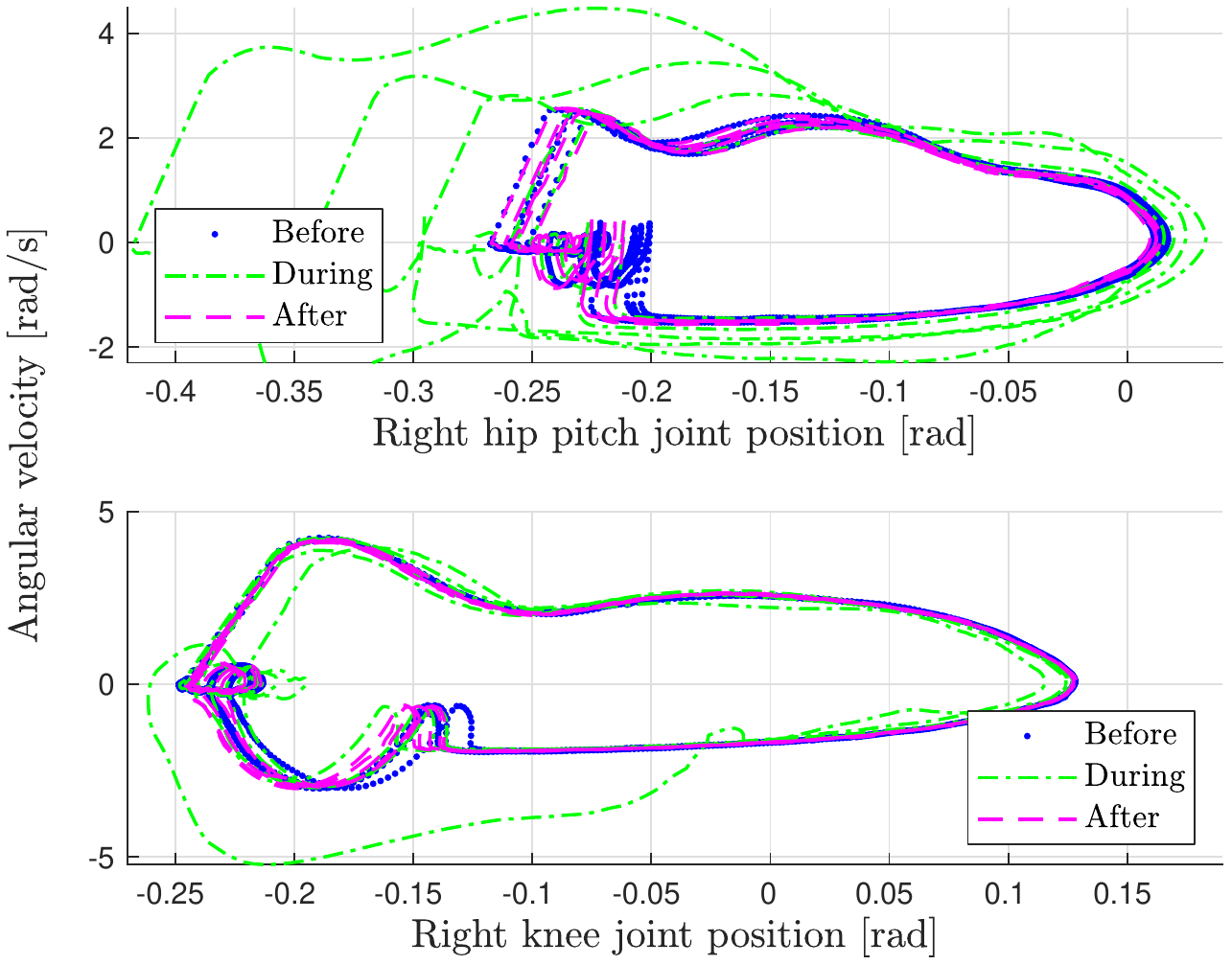}
% \vspace{-7mm}
\caption{Disturbance rejection of learned policy. The walking limit cycles described before, during and after the lateral push show that the controller can handle effectively large external disturbance forces.} 
\label{fig:disturbance}
% \vspace{-3mm}
\end{figure}

To evaluate the robustness of the policy against disturbances, adversarial forces are applied to the robot in different directions. \figref{fig:disturbancetransition} and \figref{fig:disturbance_vel} show respectively the transition of the robot recovering from a push in the lateral direction, and the velocity profile of the robot during the disturbance. In addition, \figref{fig:disturbance} shows the limit cycles of the robot's joints when the disturbance is applied, where it can be seen the robot is able to recover effectively from the push as the joint limit cycles return to a stable periodic orbit after the disturbance.

\subsection{Robustness on Different Terrains}
\begin{figure}
\centering
% \vspace{2mm}
% \includegraphics[trim={0cm 2.2cm 0cm 2.2cm},clip,width=0.9\columnwidth]{figures/Digit_v3_standing.jpg}
\includegraphics[clip,width=1\columnwidth]{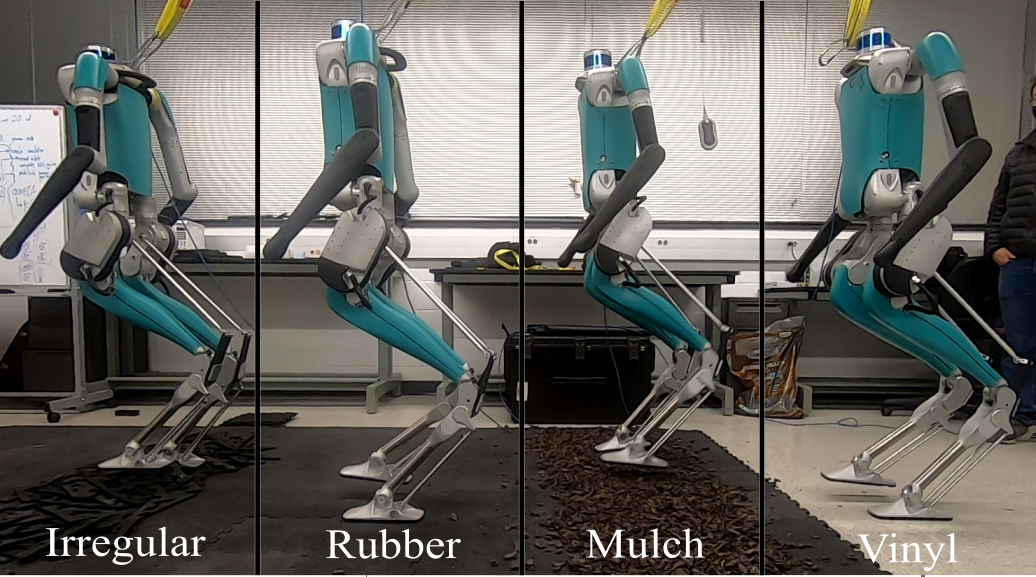}
% \vspace{-7mm}
\caption{Digit walking on different terrains. From right to left we have vinyl, mulch, flat rubber, and irregular rubber terrain. The stance foot is kept passive during the stance phase to allow the foot to adapt easily to different terrains.} 
\label{fig:terrains}
% \vspace{-3mm}
\end{figure}

To further evaluate the robustness of the learned control policy, we make Digit walk in various terrains with different levels of difficulty. These terrains include flat vinyl ground, mulch,  flat rubber ground, and irregular rubber terrain. The controller is able to adapt to any of these terrains while keeping an stable and robust walking gait. Snapshots of the walking gait over the different terrains are shown in \figref{fig:terrains}, and the complete motion can be found in the accompanying video  \cite{video_link}. %%whereas \figref{fig:foot_zoom} shows 

% \begin{figure}
% \centering
% % \vspace{2mm}
% % \includegraphics[trim={0cm 2.2cm 0cm 2.2cm},clip,width=0.9\columnwidth]{figures/Digit_v3_standing.jpg}
% \includegraphics[clip,height=3cm]{figures/foot_zoom1.png}
% \includegraphics[clip,height=3cm]{figures/foot_zoom2.png}

% % \vspace{-7mm}
% \caption{Digit walking robustly on irregular challenging terrain. The stance foot is kept passive during the stance phase to allow the foot to adapt easily to irregular and challenging terrains.  } 
% \label{fig:limit_cycle_sim}
% % \vspace{-3mm}
% \end{figure}

\section{Conclusions} \label{sec:conclusions}
In this paper, we present a framework for learning robust bipedal locomotion policies that can be transferred to real hardware with minimal tuning. By combining a sample efficient learning structure with intuitive but powerful feedback regulations in a cascade structure, we decouple the learning problem into 2 stages that work at a different frequency to facilitate the implementation of the controller in the real hardware. While the trajectory planning stage is handled by the neural network to produce reference trajectories for the actuated joints of the robot at a lower frequency (250 Hz), the feedback regulation stage runs at a higher frequency (1 kHz) using the sensor feedback to realize compensations of the reference trajectories that guarantee the stability of the walking limit cycle. The end result are policies learned from scratch that are transferred successfully to hardware with minimal tuning. The controller is exhaustively tested in hardware demonstrating stable walking gaits that are robust to external disturbances and challenging terrain without being trained under those conditions. 

% The framework implemented in this work fix the arms' joints during the walking motion to reduce the number of joints to be controlled.

% Although the framework realize robust and stable walking motion in different directions, the maximum walking speed is still limited (about 0.2 m/s). Further testing and development of the framework will look for increasing the range of walking speed and include the motion of the arms into the learning/control process. 

\bibliography{bib/IEEEabrv.bib, bib/ms.bib}
\bibliographystyle{IEEEtran}

\end{document}